\documentclass[conference]{IEEEtran}

\usepackage{cite}
\usepackage{amsmath,amssymb}
\usepackage{graphicx}
\usepackage{textcomp}
\usepackage[table]{xcolor}
\usepackage{url}
\usepackage{microtype}
\usepackage{tikz}
\usepackage[hidelinks]{hyperref}
\usepackage{float}
\usepackage{booktabs}
\usepackage{tabularx}
\usepackage{array}
\usepackage{pifont}
\usetikzlibrary{arrows.meta, positioning}

\definecolor{gmarkblue}{HTML}{EAF3FF}
\definecolor{fullgreen}{HTML}{DFF3DF}
\definecolor{partialyellow}{HTML}{FFF1CC}
\definecolor{limitedgray}{HTML}{F2F2F2}

\definecolor{fullgreentext}{HTML}{176B1A}
\definecolor{partialtext}{HTML}{9A5A00}
\definecolor{limitedtext}{HTML}{555555}

\newcolumntype{Y}{>{\centering\arraybackslash}X}
\newcolumntype{L}{>{\raggedright\arraybackslash}p{0.22\textwidth}}

\newcommand{\fullsupport}{\cellcolor{fullgreen}\textcolor{fullgreentext}{\ding{51}}}
\newcommand{\partialsupport}{\cellcolor{partialyellow}\textcolor{partialtext}{$\triangle$}}
\newcommand{\nosupport}{\cellcolor{limitedgray}\textcolor{limitedtext}{\textemdash}}

\newcommand{\legendfull}{\fcolorbox{black}{fullgreen}{\strut\textcolor{fullgreentext}{\ding{51}}}}
\newcommand{\legendpartial}{\fcolorbox{black}{partialyellow}{\strut\textcolor{partialtext}{$\triangle$}}}
\newcommand{\legendnone}{\fcolorbox{black}{limitedgray}{\strut\textcolor{limitedtext}{\textemdash}}}

\newcommand{\gmark}{\textsc{G-MARK}}

\title{\gmark{}: Grounded Multi-Agent Reasoning for Cooperative Driving via Knowledge Graphs}

\author{
    \IEEEauthorblockN{
        Bhavya Gupta\textsuperscript{1},
        Onat Gungor\textsuperscript{2},
        Tajana Rosing\textsuperscript{1}
    }
    \IEEEauthorblockA{
        \textsuperscript{1}University of California, San Diego, CA, USA \\
        \textsuperscript{2}West Virginia University, Morgantown, WV, USA \\
        \{b5gupta, tajana\}@ucsd.edu, onat.gungor@wvu.edu
    }
}

\begin{document}

\maketitle

\begin{abstract}
Autonomous driving systems must operate under partial observability, where safety-critical objects may be occluded or visible only to neighboring connected vehicles. Vehicle-to-vehicle cooperation can reduce this uncertainty, but existing cooperative driving methods often compress multi-agent evidence into latent features or hidden multimodal states. As a result, they obscure which agent observed each object, whether the object is visible to the ego vehicle, and how conflicting evidence affects downstream decisions. We propose \gmark{}, a grounded multi-agent reasoning framework that converts cooperative object-centric observations into explicit provenance-aware knowledge graphs (KGs). The resulting KGs preserve object hypotheses together with their source attribution, ego-versus-partner visibility, uncertainty, conflicts, spatial relations, and planning-relevant context. \gmark{} then derives a shared feature representation from these KGs, enabling lightweight task heads to support object reasoning, motion prediction, control selection, and trajectory forecasting. Compared with the state-of-the-art baseline, \gmark{} improves occlusion reasoning accuracy by 42.2\%, reduces control-selection error by 13.1\%, and achieves comparable trajectory-planning accuracy with a 25.6$\times$ smaller structured communication payload. Our code is available at
\href{https://github.com/bhavyagupta98/g-mark}{this repository}.


\end{abstract}


\section{Introduction}
\label{sec:intro}

Autonomous driving systems must make safety-critical decisions from incomplete and viewpoint-dependent observations~\cite{zhang2021safe}. Objects that are crucial for planning may be occluded, outside the ego vehicle's field of view, or observable only by a connected partner vehicle~\cite{zhang2023occlusion}. Vehicle-to-vehicle cooperation can mitigate this partial observability by sharing complementary scene evidence across agents~\cite{xu2022opv2v}. However, when the ego vehicle must reason about an object it has not directly observed, the system should retain more than the object hypothesis itself. It should also preserve its provenance: which agent observed the object, how reliable the observation is, and why the object matters for downstream planning~\cite{kuznietsov2024explainable}.

Existing cooperative perception methods improve autonomous driving perception through early fusion, late fusion, and feature-level fusion~\cite{where2comm,cobevt,v2xvit}. These methods primarily aim to aggregate multi-agent observations into more accurate perception outputs, often by compressing shared evidence into latent feature maps or final detections. However, this compression can discard the evidence structure needed for downstream reasoning. Hence, downstream modules cannot infer which agent supported an object hypothesis, whether the ego vehicle directly observed the object, or how much uncertainty remains in the object estimate.

Recent language-based and vision-language driving systems study high-level driving reasoning through visual question answering, explanation generation, behavior planning, and language-conditioned control~\cite{drivelm,drivegpt4,lmdrive,drivevlm}. Cooperative driving systems such as V2V-LLM~\cite{chiu2026v2vllm} and V2V-GoT~\cite{chiu2026v2vgot} extend this paradigm to multi-agent reasoning. Yet, their intermediate states are often encoded as prompts, generated text, free-form rationales, or hidden multimodal features rather than structured object-level cooperative evidence. This limits traceability for safety-critical decisions that depend on objects outside the ego vehicle's direct view but observed by a connected partner.

\begin{table*}[t]
\centering
\caption{
Comparison of cooperative evidence exposed by related method families.
}
\label{tab:related_work}
\small
\setlength{\tabcolsep}{3pt}
\renewcommand{\arraystretch}{1.25}
\begin{tabularx}{\textwidth}{L Y Y Y Y Y}
\toprule
\textbf{Capability}
& \textbf{Cooperative Perception\cite{cobevt} / V2X Fusion\cite{v2xvit}}
& \textbf{Language / VLM Driving Reasoning\cite{drivelm}}
& \textbf{Structured Scene and KG-based Reasoning~\cite{tian2026kldrive}}
& \textbf{Cooperative Language-Based Reasoning~\cite{chiu2026v2vgot}}
& \cellcolor{gmarkblue}\textbf{\gmark{} (Ours)} \\
\midrule

Multi-agent evidence
& \fullsupport
& \partialsupport
& \nosupport
& \fullsupport
& \fullsupport \\

Object provenance
& \nosupport
& \nosupport
& \partialsupport
& \partialsupport
& \fullsupport \\

Ego-versus-partner visibility
& \nosupport
& \partialsupport
& \nosupport
& \partialsupport
& \fullsupport \\

Uncertainty / conflict
& \nosupport
& \partialsupport
& \partialsupport
& \partialsupport
& \fullsupport \\

Planning relevance
& \partialsupport
& \fullsupport
& \partialsupport
& \fullsupport
& \fullsupport \\

Reusable across tasks
& \nosupport
& \fullsupport
& \partialsupport
& \fullsupport
& \fullsupport \\

Comm. efficiency
& \partialsupport
& \partialsupport
& \nosupport
& \partialsupport
& \fullsupport \\

\bottomrule
\end{tabularx}

\vspace{0.45em}
\begin{center}
\small
\legendfull~Explicitly modeled
\hspace{1.5em}
\legendpartial~Partially exposed
\hspace{1.5em}
\legendnone~Not explicitly exposed
\end{center}
\end{table*}

We propose \gmark{}, a grounded multi-agent reasoning framework that represents cooperative driving scenes as provenance-aware knowledge graphs. Operating above the perception layer, \gmark{} consumes processed multi-agent observations, associates compatible observations into shared object hypotheses, and preserves weak or unmatched observations as uncertain candidates rather than discarding them. The resulting graph explicitly stores source-agent support, ego-versus-partner visibility, uncertainty, cross-agent disagreement, spatial relations, motion cues, and planning context. KG-based task heads then query this shared KG to support object grounding, occlusion reasoning, hidden-object discovery, motion prediction, control selection, and trajectory forecasting.

We evaluate \gmark{} on V2V-GoT-QA, a cooperative driving benchmark spanning perception, prediction, and planning tasks. 
Compared with V2V-GoT~\cite{chiu2026v2vgot}, \gmark{} improves occlusion reasoning by $42.2\%$, hidden-object discovery by $12.3\%$, and control-selection error by $13.1\%$. 
For future trajectory forecasting, \gmark{} reaches near-parity in average L2 error while using approximately $25.6\times$ lower structured communication. Our contributions are summarized as follows:
\begin{itemize}
    \item We introduce a provenance-aware cooperative KG that preserves object hypotheses, source-agent support, visibility, uncertainty, disagreement, spatial relations, motion cues, and planning context.
    \item We propose task-conditioned delayed evidence fusion, which retains weak partner-only observations so downstream tasks can decide which evidence is relevant.
    \item We show that our approach can support object reasoning, motion prediction, control selection, and future trajectory forecasting with compact structured communication.
\end{itemize}

\section{Related Work}
\label{sec:related}

Prior work relevant to \gmark{} spans cooperative perception and V2X fusion~\cite{where2comm,v2xvit,cobevt}, language- and VLM-based driving reasoning~\cite{drivelm,drivevlm,lmdrive,drivegpt4}, structured scene and KG-based reasoning~\cite{hogan2021knowledge,tian2026kldrive}, and cooperative language-based reasoning~\cite{chiu2026v2vllm,chiu2026v2vgot,langcoop,colmdriver}. \gmark{} targets the gap between these research families by constructing an explicit cooperative evidence representation from perception outputs. This representation preserves source support, ego-versus-partner visibility, uncertainty, disagreement, and planning context before downstream task inference. These properties are important because cooperative driving decisions depend not only on which objects are present, but also on who observed them, whether they are visible to the ego vehicle, and why they affect planning~\cite{kuznietsov2024explainable}. Table~\ref{tab:related_work} compares how existing method families expose these evidence variables for downstream reasoning.

\textbf{Cooperative Perception and V2X Fusion.}
Cooperative perception improves scene understanding beyond a single ego viewpoint by sharing raw observations, object predictions, learned features, BEV representations, or sparse spatial messages across agents. 
Representative systems include Where2Comm~\cite{where2comm}, V2X-ViT~\cite{v2xvit}, and CoBEVT~\cite{cobevt}. 
These methods provide strong perception backbones, but their fused outputs rarely expose source support, ego-versus-partner visibility, uncertainty, or disagreement as reusable object-level evidence. 

\textbf{Language and VLM-based Driving Reasoning.}
Language, vision-language, and multimodal large language models have recently been used for driving question answering, explanation, planning, and control~\cite{drivelm,drivegpt4,lmdrive,drivevlm}. 
These methods broaden autonomous-driving evaluation from perception accuracy to reasoning-oriented tasks, including object grounding, scene understanding, decision explanation, and planning. 
However, most are ego-centric or non-cooperative, motivating a layer that preserves source-agent support and cross-agent disagreement.

\textbf{Structured Scene and KG-based Reasoning.}
Structured scene representations and knowledge graphs encode entities, attributes, and relations in an explicit form for visual reasoning and decision making~\cite{hogan2021knowledge}. 
In autonomous driving, systems such as DriveLM~\cite{drivelm} and KLDrive~\cite{tian2026kldrive} show that structured scene facts can support interpretable, planning-oriented reasoning. 
However, existing driving KG approaches primarily focus on ego-centric scenes, language-grounded QA, or high-level facts rather than cooperative V2V evidence.
\gmark{} extends this direction with a per-scene cooperative KG that stores object hypotheses, source-agent support, visibility, uncertainty, disagreement, and planning relevance.

\textbf{Cooperative Language-Based Reasoning.}
Recent cooperative language-based methods extend language reasoning to multi-agent driving. 
LangCoop uses language for compact collaborative communication~\cite{langcoop}, CoLMDriver studies LLM-based negotiation~\cite{colmdriver}, V2V-LLM formulates vehicle-to-vehicle reasoning as multimodal QA~\cite{chiu2026v2vllm}, and V2V-GoT organizes perception, prediction, and planning questions through graph-of-thought reasoning~\cite{chiu2026v2vgot}. 
These systems are closest to \gmark{} in task scope, but their intermediate states are typically language messages, generated answers, reasoning traces, or hidden multimodal activations rather than explicit object-level evidence.
\gmark{} instead uses a provenance-aware KG, allowing downstream predictions to be traced to source agents, visibility facts, uncertainty, and supporting observations.

\section{\gmark{} Framework}
\label{sec:framework}

\begin{figure*}[t]
    \centering
    \includegraphics[width=.95\textwidth]{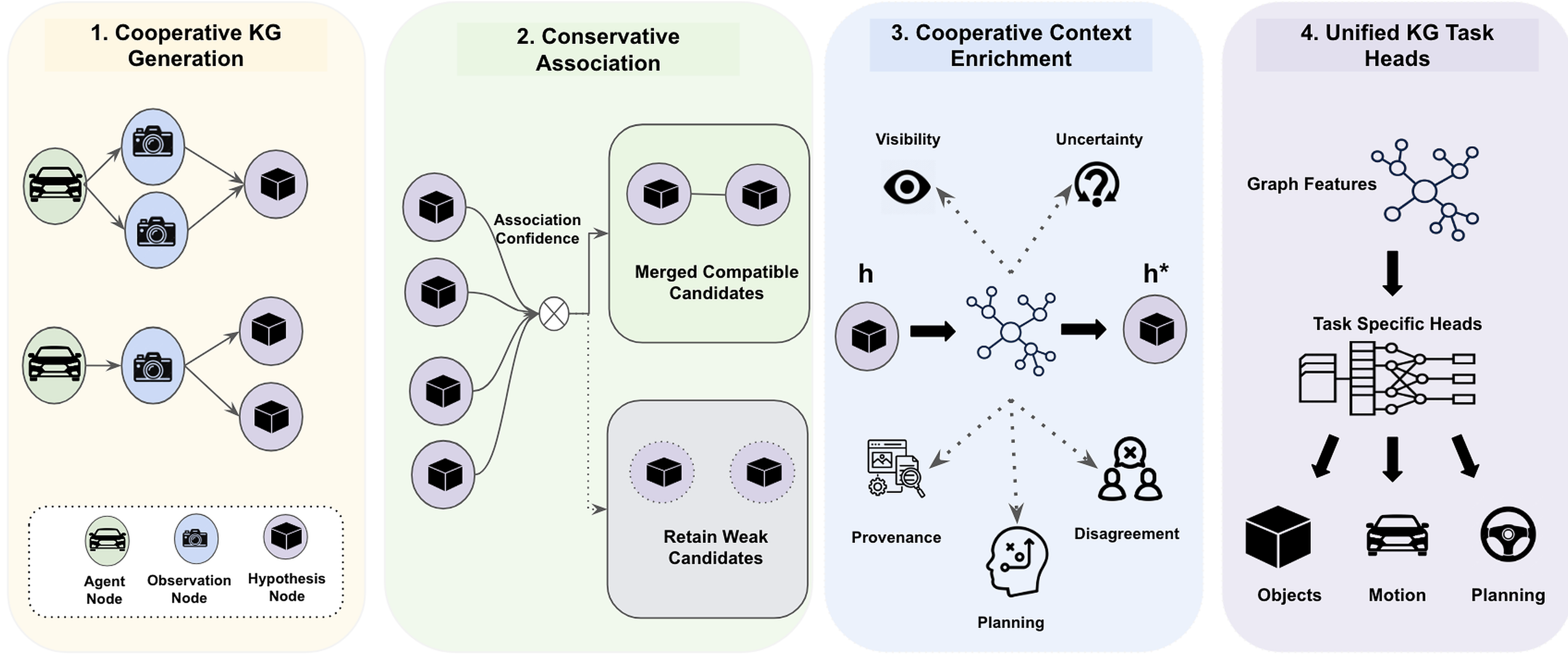}
    \caption{
        Overview of the \gmark{} cooperative KG reasoning framework.
    }
    \label{fig:gmark_overview}
\end{figure*}


Fig.~\ref{fig:gmark_overview} provides an overview of \gmark{}, a cooperative reasoning layer that operates on top of a cooperative perception pipeline. Given processed multi-agent scene artifacts, including object detections or tracks, agent poses, confidence scores, and planning context, \gmark{} constructs a provenance-aware KG for downstream reasoning. Importantly, \gmark{} does not consume raw camera or LiDAR streams directly; instead, it abstracts already-processed cooperative perception evidence into a structured graph representation. The central design principle of \gmark{} is to postpone irreversible fusion until the downstream reasoning task is specified. Rather than merging all agent observations into a single fused object list, \gmark{} represents observing agents, local observations, and object hypotheses as linked graph entities. This structure allows different reasoning heads, such as occlusion reasoning and motion prediction, to query the same cooperative evidence while emphasizing different task-relevant factors.
 
\gmark{} proceeds in four stages. 
First, each CAV's processed perception output is instantiated as a local evidence KG containing an agent node, observation nodes, and object-hypothesis nodes before cross-agent association. 
Second, ego-side conservative association links compatible local evidence across CAVs into shared object hypotheses, attaches observation nodes as support, and retains weak unmatched candidates. 
Third, each hypothesis is enriched with provenance, agent-relative visibility, uncertainty, disagreement, spatial relations, motion cues, and planning relevance. 
Finally, KG-based task heads query the enriched cooperative KG for object-level reasoning, motion prediction, and trajectory planning.

\subsection{Cooperative KG Generation}

\gmark{} represents each CAV's processed perception output, including local observations and V2V-communicated object hypotheses, as local KG evidence and assembles it into an initial cooperative KG, instead of collapsing it into a flat list of fused objects~\cite{hogan2021knowledge,tian2026kldrive}. A flat object list can indicate that an object exists, but it does not preserve who observed it, whether it is ego-visible or partner-only, how reliable the supporting evidence is, or whether the object should influence planning. These factors are especially important under occlusion and partial observability, where safety-relevant evidence may be available only through another connected agent~\cite{v2v4real,v2xsim,v2xvit}.

To preserve the evidence structure behind cooperative perception outputs, each local evidence graph separates \textit{agent nodes}, \textit{observation nodes}, and \textit{object-hypothesis nodes}. 
An \textit{agent node} stores the identity, pose, timestamp, and ego-versus-partner role of the CAV that produced the local evidence. 
An \textit{observation node} represents one agent-local object observation $o^{(i)}_{t,j}$, while an \textit{object-hypothesis node} represents an object candidate from communicated CAV evidence or a shared hypothesis formed after cross-agent association. This separation preserves the source vehicle, supporting observation, and ego-versus-partner visibility of each hypothesis. 
For example, if a partner observes a pedestrian hidden from the ego vehicle, the agent node identifies the partner, the observation node stores the local detection, and the hypothesis node represents the shared belief that the pedestrian exists.

Let $\mathcal{A}_t=\{a_1,\ldots,a_{N_t}\}$ denote the set of connected agents available at timestamp $t$, where $N_t$ is the number of agents at that timestamp. 
Each agent $a_i \in \mathcal{A}_t$ reports a set of $M_{i,t}$ local object observations:
\[
    \mathcal{O}^{(i)}_t
    =
    \{o^{(i)}_{t,1},\ldots,o^{(i)}_{t,M_{i,t}}\}.
\]
Here, $M_{i,t}$ is the number of local observations from agent $a_i$ at time $t$, and $j=1,\ldots,M_{i,t}$ indexes them.
Each observation is represented as
\[
    o^{(i)}_{t,j}
    =
    \left(
    \mathbf{x}^{(i)}_{t,j},
    \tau^{(i)}_{t,j},
    c^{(i)}_{t,j},
    a_i,
    t,
    \boldsymbol{\eta}^{(i)}_{t,j}
    \right),
\]
where $j$ denotes an individual observation, $\mathbf{x}^{(i)}_{t,j}$ is the object position or bounding box in a shared coordinate frame, $\tau^{(i)}_{t,j}$ is the semantic object type, such as car, truck, or pedestrian, $c^{(i)}_{t,j}$ is the detection confidence, and $a_i$ records the source agent. 
The auxiliary vector $\boldsymbol{\eta}^{(i)}_{t,j}$ stores optional observation-level attributes, such as velocity, track identity, source flags, or visibility labels. Graph-level quantities such as hypothesis-level provenance, uncertainty, disagreement, and planning relevance are finalized after association.
Each local observation $o^{(i)}_{t,j}$ is instantiated as an observation node $v^{O}_{t,i,j} \in \mathcal{V}^{O}_t$. 
One agent node $v^{A}_{t,i} \in \mathcal{V}^{A}_t$ is created for each connected vehicle $a_i \in \mathcal{A}_t$. 
Each observation node $v^{O}_{t,i,j}$ is linked to the agent node $v^{A}_{t,i}$ that generated it, making the source of every observation explicit in the KG.

The CAV-local evidence is then assembled into an initial cooperative KG:
\[
    \mathcal{G}_t = (\mathcal{V}_t,\mathcal{E}_t),
\]
with node set
\[
    \mathcal{V}_t =
    \mathcal{V}^{A}_t
    \cup
    \mathcal{V}^{O}_t
    \cup
    \mathcal{V}^{H}_t .
\]
Here, $\mathcal{V}^{A}_t$, $\mathcal{V}^{O}_t$, and $\mathcal{V}^{H}_t$ denote agent nodes, observation nodes, and object-hypothesI removedis nodes, respectively.

The relation set $\mathcal{E}_t$ uses typed relations for source attribution, hypothesis support, spatial context, planning context, and evidence consistency. 
The source relation connects an agent node to each observation node it produces, while the support relation connects an observation node to the object-hypothesis node it contributes to. 
Spatial relations, such as \textit{front-of}, \textit{behind}, \textit{left-of}, \textit{right-of}, and \textit{near}, are computed in the ego coordinate frame. 
Planning-context relations, such as \textit{near-trajectory} and \textit{path-relevant}, use distances to the ego vehicle's planned or predicted trajectory. 
Evidence-consistency relations include \textit{cooperatively-supported}, indicating support from multiple agents or from a partner agent, and \textit{low-conflict}, indicating geometrically consistent support. 
This typed relation design follows standard entity--relation KG representations~\cite{hogan2021knowledge}, while provenance is stored explicitly through the support set and provenance record~\cite{w3cprov}.

An object-hypothesis node $h \in \mathcal{V}^{H}_t$ is associated with a support set of compatible observation nodes,
\[
    \mathcal{S}(h) \subseteq \mathcal{V}^{O}_t.
\]
Thus, $\mathcal{S}(h)$ contains observation nodes, each corresponding to an original local observation $o^{(i)}_{t,j}$. It is represented as
\[
    h =
    (\mathbf{x}_h, \tau_h, c_h, \pi_h, \mathcal{S}(h), u_h, \gamma_h, r_h),
\]
where $(\mathbf{x}_h,\tau_h,c_h)$ denote fused location, semantic type, and confidence; $(\pi_h,\mathcal{S}(h))$ store supporting agents and observations; and $(u_h,\gamma_h,r_h)$ capture uncertainty, disagreement, and planning relevance.

The provenance record is
\[
    \pi_h=\{(a_i,v^{O}_{t,i,j}) : v^{O}_{t,i,j} \in \mathcal{S}(h),\; v^{O}_{t,i,j} \text{ is produced by } a_i\}.
\]
This record keeps each object hypothesis traceable to its supporting agents and observations~\cite{w3cprov}.
The KG exposes cooperative evidence variables that are often lost after immediate fusion. Source support is computed from the number and identity of agents in $\pi_h$. Ego-relative visibility is represented as an agent-specific relation,
\[
    \mathrm{vis}(h,a_i) \in
    \{\mathrm{visible},\mathrm{occluded},\mathrm{uncertain}\},
\]
rather than a global label. 
Downstream heads use this relation to distinguish ego-visible hypotheses from partner-only or occluded hypotheses, especially for invisible-object retrieval, occlusion reasoning, and planning-aware object selection. 
Uncertainty $u_h$ captures weak confidence, candidate status, and limited support; disagreement $\gamma_h$ captures inconsistent observations; and planning relevance $r_h$ measures proximity to the ego vehicle's future path.

The output of this stage is an initial cooperative KG $\mathcal{G}_t$ containing agent nodes, observation nodes, object-hypothesis nodes, typed relations, and traceable evidence variables. 
The following stages resolve compatible candidate hypotheses through conservative association and enrich each hypothesis with provenance, visibility, uncertainty, disagreement, and planning relevance.

\subsection{Conservative Association}

After local evidence is assembled into the cooperative KG, \gmark{} associates compatible candidate hypotheses across CAVs into shared cooperative object hypotheses. 
Observation nodes are attached as hypothesis support, while unmatched observations are retained as weak candidates.
This avoids over-merging nearby but distinct objects while preserving partner-only evidence for occlusion and planning.

Following standard tracking and data-association practice, we use a deterministic gating rule to decide which local evidence pairs are eligible for association~\cite{barshalom1988tracking}. 
For observation-level support attachment, let $v_p^O,v_q^O \in \mathcal{V}^{O}_t$ denote two observation nodes, with semantic types $\tau_p,\tau_q$ and positions $\mathbf{x}_p,\mathbf{x}_q$ inherited from their corresponding local observations. 
We define the association gate as
\[
    \delta_t(v_p^O,v_q^O)
    =
    \mathbf{1}
    \left[
        \mathrm{type\_match}(\tau_p,\tau_q)
        \land
        d(\mathbf{x}_p,\mathbf{x}_q) < \epsilon_{\tau_p,\tau_q}
    \right],
\]
where $d(\cdot,\cdot)$ is the distance between observation centers and $\epsilon_{\tau_p,\tau_q}$ is a class-dependent association threshold. 
The function $\mathrm{type\_match}(\tau_p,\tau_q)$ returns true only when the two observations have the same semantic class. We use exact semantic-class matching as a conservative design choice because false cross-class merges can create unsafe object hypotheses. For example, nearby vehicle observations can be associated, but a pedestrian and a vehicle remain separate even if they are geometrically close.

For gated pairs, \gmark{} assigns an interpretable distance-based affinity score,
\[
    \alpha_{t,pq}
    =
    \begin{cases}
    1-\dfrac{d(\mathbf{x}_p,\mathbf{x}_q)}{\epsilon_{\tau_p,\tau_q}},
    & \text{if } \delta_t(v_p^O,v_q^O)=1,\\
    0,
    & \text{otherwise.}
    \end{cases}
\]
The affinity score ranks gated pairs by spatial agreement, with higher values assigned to closer pairs. 
The gating criterion follows standard data-association practice~\cite{barshalom1988tracking}, while the normalized distance score is an interpretable heuristic used in \gmark{}. When an unmatched observation node is promoted to a candidate hypothesis, its representative state inherits the observation's semantic type and position, so the same semantic and geometric compatibility rule is used for later candidate-hypothesis merging. The association graph induced by $\delta_t(v_p^O,v_q^O)=1$ determines which observation nodes attach as support evidence. 
For candidate hypotheses, the same semantic and geometric rule is applied to their representative states to identify candidates eligible for merging. 
The fused position $\mathbf{x}_h$ is computed as a confidence-weighted average of the observation positions in $\mathcal{S}(h)$.

For each object hypothesis, \gmark{} summarizes confidence using a bounded noisy-or-style aggregation~\cite{koller2009probabilistic}:
\[
    c_h
    =
    1 - \prod_{v^O \in \mathcal{S}(h)} (1-c(v^O)),
\]
where $c(v^O)$ is the confidence stored by observation node $v^O$. 
We use $c_h$ as a bounded confidence summary rather than as a calibrated probabilistic fusion model. Unlike immediate fusion, the association output is not only a fused object location. 
For every hypothesis, \gmark{} stores the support set $\mathcal{S}(h)$, provenance record $\pi_h$, confidence summary $c_h$, candidate status, uncertainty, and disagreement attributes. 
If an observation node does not pass the association gate with any other observation node, \gmark{} retains it as a candidate hypothesis with lower support and higher uncertainty. 
Candidate retention is a safety-oriented design choice: partner-only or partially observed objects remain available for downstream inference, but are not treated as fully reliable detections.

This module outputs well-supported hypotheses with attached observation evidence, merged compatible candidates, and retained weak candidates.
The next module enriches them with provenance, visibility, uncertainty, disagreement, and planning relevance.

\subsection{Cooperative Context Enrichment}

After hypothesis formation, \gmark{} derives context features for each object hypothesis from support, visibility, relative geometry, and ego motion context. 
Unlike association, this stage describes how the belief was formed, who can observe it, how reliable it is, whether agents disagree, and whether it is relevant to ego motion.
We group these enrichments into five categories, summarized in Table~\ref{tab:kg_enrichment}. Together, these enrichments let downstream modules reason over object state, supporting evidence, and ego-motion relevance~\cite{drivelm}.

\begin{table}[t]
\centering
\caption{Cooperative context enrichment categories in \gmark{}.}
\label{tab:kg_enrichment}
\small
\begin{tabular}{>{\raggedright\arraybackslash}p{0.33\linewidth} >{\raggedright\arraybackslash}p{0.52\linewidth}}
\toprule
\textbf{Category} & \textbf{Role in downstream reasoning} \\
\midrule
Provenance and support
& Stores source agents, support counts, and ego/partner/cooperative status to trace object evidence. \\

Agent visibility
& Stores ego-relative and agent-relative visibility for occlusion and hidden-object reasoning. \\

Uncertainty and reliability
& Marks weak, ambiguous, or candidate hypotheses so they are not treated as fully reliable. \\

Cross-agent disagreement
& Records geometric or semantic inconsistency among agents to expose conflicting evidence. \\

Planning relevance
& Stores trajectory proximity, motion cues, and control context. \\
\bottomrule
\end{tabular}
\end{table}

These enrichments are needed because the same object can matter differently across tasks: partner-only objects support hidden-object discovery, uncertainty affects prediction confidence, and path proximity affects control and forecasting. Formally, for each object hypothesis $h$, \gmark{} augments the node with an enrichment vector
\[
    \phi(h) =
    [
    \phi_{\mathrm{prov}}(h),
    \phi_{\mathrm{vis}}(h),
    \phi_{\mathrm{unc}}(h),
    \phi_{\mathrm{dis}}(h),
    \phi_{\mathrm{plan}}(h)
    ],
\]
where $\phi_{\mathrm{prov}}(h)$, $\phi_{\mathrm{unc}}(h)$, and $\phi_{\mathrm{dis}}(h)$ summarize support, provenance, confidence, and association consistency from the conservative association stage.
The components $\phi_{\mathrm{vis}}(h)$ and $\phi_{\mathrm{plan}}(h)$ encode ego-relative visibility and distance to the ego vehicle's planned or predicted trajectory. The final KG-derived feature vector $\psi(h)$ concatenates object-state attributes, such as position, semantic type, confidence, and motion cues, with the enrichment vector $\phi(h)$. 

The output is an enriched cooperative KG in which downstream modules can query each hypothesis through $\psi(h)$ without recomputing evidence from raw observations.

\begin{table}[t]
\centering
\caption{KG-based task-head families in \gmark{}.}
\label{tab:task_heads}
\small
\begin{tabular}{>{\raggedright\arraybackslash}p{0.33\linewidth} >{\raggedright\arraybackslash}p{0.52\linewidth}}
\toprule
\textbf{Head family} & \textbf{Role in downstream reasoning} \\
\midrule
Selection and visibility reasoning
& Retrieves visible, occluded, hidden, or planning-relevant objects using KG evidence. \\

Motion and interaction prediction
& Predicts object or agent behavior from state, motion, support, uncertainty, and path-relative geometry. \\

Control and trajectory planning
& Predicts controls or future waypoints from risk, planning relevance, geometry, and trajectory context. \\
\bottomrule
\end{tabular}
\end{table}

\begin{table*}[t]
\centering
\caption{Main results on V2V-GoT-QA. \gmark{} improves most strongly on tasks that require explicit cooperative evidence, including occlusion reasoning and hidden-object discovery.}
\label{tab:main_v2vgotqa_results}
\small
\setlength{\tabcolsep}{4pt}
\renewcommand{\arraystretch}{1.12}
\begin{tabularx}{\textwidth}{@{}l X l c c c@{}}
\toprule
\textbf{Task} & \textbf{Task family} & \textbf{Metric} & \textbf{\gmark{}} & \textbf{V2V-GoT~\cite{chiu2026v2vgot}} & \textbf{Improvement} \\
\midrule
Notable Objects (Q1) & Visible-object grounding & F1@0.5m $\uparrow$ & \textbf{0.586} & 0.525 & +11.6\% \\
Occluding Objects (Q2) & Occlusion reasoning & F1@0.5m $\uparrow$ & \textbf{0.428} & 0.301 & +42.2\% \\
Invisible Objects (Q3) & Hidden-object discovery & F1@0.5m $\uparrow$ & \textbf{0.494} & 0.440 & +12.3\% \\
Planning Awareness (Q4) & Planning-relevance reasoning & F1@0.5m $\uparrow$ & \textbf{0.614} & 0.608 & +0.9\% \\
Object Motion (Q5) & Object motion prediction & L2 avg. $\downarrow$ & \textbf{3.822} & 8.050 & \textbf{+52.5\%} \\
Agent Motion (Q6) & Agent motion prediction & Accuracy $\uparrow$ & \textbf{0.905} & 0.874 & +3.5\% \\
Object Motion (Q7) & Object motion prediction & L2 avg. $\downarrow$ & \textbf{3.822} & 7.610 & \underline{+49.8\%} \\
Control Settings (Q8) & Control selection & Action L1 $\downarrow$ & \textbf{0.076} & 0.088 & +13.1\% \\
Future Trajectory (Q9) & Future trajectory forecasting & L2 avg. $\downarrow$ & 2.710 & \textbf{2.620} & -3.4\% \\
\bottomrule
\end{tabularx}
\end{table*}

\subsection{Unified KG Task Heads}

The enriched KG provides a shared evidence state for task heads with different output formats. 
As shown in Table~\ref{tab:task_heads}, we group them into object selection and visibility reasoning, motion and interaction prediction, and control and trajectory planning, following the standard perception-prediction-planning decomposition~\cite{yurtsever2020survey}. 
Each head operates on KG-derived features $\psi(h)$ that combine object state, provenance, visibility, uncertainty, disagreement, spatial relations, and planning-context attributes.

For a task family $k$, we write the inference head as
\[
    y_k = R_k\!\left(\mathcal{G}_t,\{\psi(h):h\in\mathcal{V}^{H}_t\}\right),
\]
where $\mathcal{G}_t$ is the enriched cooperative KG, $\psi(h)$ is the KG-derived feature vector for hypothesis $h$, $R_k$ is the task head, and $y_k$ is the structured task output. 
Depending on the task, $R_k$ is implemented as either a deterministic retrieval/ranking function or a lightweight learned predictor over $\psi(h)$. 
Learned heads support motion regression, scene-level prediction, control selection, and trajectory forecasting without heavy multimodal inference, while preserving evidence traceability.
Table~\ref{tab:task_heads} summarizes how the same enriched KG supports different task families through a common feature interface. 
This design decouples representation construction from task inference: \gmark{} builds one provenance-aware cooperative KG per scene, and each head selects the evidence needed for its output. 
In our benchmark instantiation, these heads cover object grounding, visibility and occlusion reasoning, planning-aware object selection, motion prediction, control selection, and future trajectory forecasting.

\section{Experimental Analysis}
\label{sec:exps}

\subsection{Experimental Setup}

\textbf{Benchmarks.}
V2V4Real provides synchronized real-world V2V scenes with connected vehicles, object annotations, agent poses, timestamps, and processed perception artifacts. 
V2V-GoT-QA extends this setting with typed object-selection, visibility-reasoning, motion-prediction, control-selection, and trajectory-forecasting tasks. 
We use the official splits, with approximately $110$K training and $31$K validation questions, and evaluate on the validation split using the same task definitions and reference answers as V2V-GoT~\cite{chiu2026v2vgot}.

\textbf{Evaluation metrics.}
Object selection and visibility reasoning tasks use localization-aware precision, recall, and F1-score, where predictions must match ground-truth coordinates within the benchmark threshold. 
Motion and trajectory outputs use L2 error, agent motion uses binary accuracy, and control selection uses normalized action error. 
Higher is better for precision, recall, F1, and accuracy; lower is better for L2 and action error.

\textbf{Baselines.}
Our primary baseline is V2V-GoT~\cite{chiu2026v2vgot}, the closest recent cooperative reasoning method for this benchmark. V2V-GoT extends V2V-LLM-style cooperative multimodal QA~\cite{chiu2026v2vllm} by incorporating graph-of-thought reasoning for perception, prediction, and planning questions. It reports strong performance on future trajectory forecasting and other high-level cooperative driving tasks. We therefore compare \gmark{} with the reported V2V-GoT results across all benchmark tasks, including object reasoning, motion prediction, control selection, and trajectory forecasting.

Communication efficiency is an important requirement for cooperative driving, where agents must share sufficient scene evidence without incurring excessive bandwidth overhead. For the communication--accuracy analysis, we compare \gmark{} with the future-trajectory forecasting results of No Fusion, AttFuse~\cite{xu2022opv2v}, V2X-ViT~\cite{v2xvit}, CoBEVT~\cite{cobevt}, V2V-LLM~\cite{chiu2026v2vllm}, and V2V-GoT~\cite{chiu2026v2vgot}. We restrict this analysis to future trajectory forecasting because this task directly links cooperative communication cost to downstream planning accuracy.

These comparisons evaluate whether explicit cooperative evidence improves visibility-sensitive reasoning, generalizes across multiple task families, and offers a communication-efficient alternative to language-mediated cooperative reasoning.

\textbf{Implementation details.}
All experiments use the KG construction pipeline over processed cooperative perception artifacts, including object boxes, confidence scores, agent poses, trajectory context, and scene metadata. After KG construction, \gmark{} exports a shared KG-derived feature bank with three typed views: object retrieval for Q1--Q4, motion regression for Q5/Q7/Q9, and scene-action prediction for Q6/Q8.

Object retrieval uses a shared logistic regression ranker with task one-hot features and thresholds selected from the training split. 
Motion prediction uses regularized regression heads, with a shared Q5/Q7 object-motion head and a separate Q9 trajectory head. 
Action prediction uses lightweight classifiers for Q6 and separate speed and steering heads for Q8. 
All heads are trained on the official training split, with validation labels used only for evaluation; experiments run on a Linux machine with 8 CPU cores and 24 GB RAM.

\subsection{Results}

\textbf{Explicit cooperative evidence improves visibility-sensitive reasoning.}
Table~\ref{tab:main_v2vgotqa_results} compares \gmark{} with V2V-GoT~\cite{chiu2026v2vgot}, a state-of-the-art cooperative reasoning baseline for V2V-GoT-QA, across all task categories. 
We report Object Motion Q5 and Q7 separately because they are distinct benchmark entries, although they share the same output family. The strong gains on these entries likely reflect the structure of the output: object-motion prediction is a numeric regression problem, and \gmark{} directly exposes object state, recent motion cues, and path-relative geometry as compact KG-derived features. 
This representation is well suited to motion regression because $\psi(h)$ directly encodes numeric object state, recent motion cues, and path-relative geometry.
Language-mediated reasoning baselines must infer these quantities less directly before producing a numeric output.

\gmark{} improves over V2V-GoT on eight of the nine benchmark entries, with the largest gains on tasks that require explicit cooperative evidence. 
Occlusion reasoning improves by $42.2\%$, hidden-object discovery by $12.3\%$, and object motion prediction by $49.8\%$. 
These gains support task-conditioned delayed evidence fusion: preserving provenance, visibility, uncertainty, disagreement, and candidate status helps each task head query relevant evidence rather than relying on a single collapsed object state. 
The shared KG also supports prediction and control, with object motion benefiting from KG-derived motion cues and path-relative geometry, and control-selection error reduced by $13.1\%$. Future trajectory forecasting shows a different pattern: \gmark{} reaches near-parity with V2V-GoT, with an average L2 error of $2.710$m compared with $2.620$m. 
This suggests that the KG-conditioned representation captures useful planning context, but longer-horizon forecasting remains sensitive to accumulated speed, steering, and curvature errors. 

\begin{figure}[t]
    \centering
    \includegraphics[width=.95\columnwidth]{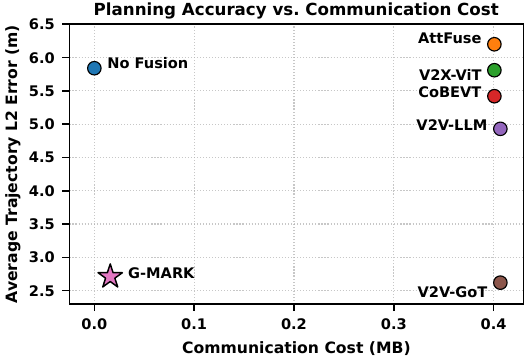}
    \caption{Planning accuracy versus communication cost.}
    \label{fig:comm_l2_tradeoff}
\end{figure}

\textbf{\gmark{} approaches planning accuracy with substantially lower communication.}
We use future trajectory forecasting because it tests whether communication preserves planning-relevant evidence, including nearby objects, occluded or partner-only hazards, and path-relative interactions. 
Trajectory error therefore measures useful scene information retained per unit of communication.
Fig.~\ref{fig:comm_l2_tradeoff} compares \gmark{} with three categories of cooperative planning baselines: a non-cooperative baseline (No Fusion), intermediate-fusion baselines including AttFuse~\cite{xu2022opv2v}, V2X-ViT~\cite{v2xvit}, and CoBEVT~\cite{cobevt}, and language-mediated cooperative reasoning baselines including V2V-LLM~\cite{chiu2026v2vllm} and V2V-GoT~\cite{chiu2026v2vgot}. No Fusion has high trajectory error, while intermediate-fusion and language-mediated methods improve accuracy at roughly $0.4008$--$0.4068$ MB per sample. 
\gmark{} achieves a different operating point: it uses only $0.0159$ MB per sample, giving about $25.2\times$ lower communication than intermediate-fusion methods and $25.6\times$ lower communication than V2V-GoT. 
This reduction comes from transmitting compact KG evidence rather than dense feature-level or language-mediated payloads. Although V2V-GoT achieves the lowest trajectory error, \gmark{} attains comparable forecasting accuracy with far lower communication and explicit evidence traceability.

\begin{figure}[t]
    \centering
    \includegraphics[width=.95\columnwidth]{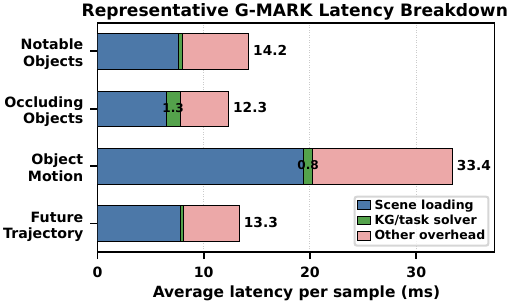}
    \caption{End-to-end latency breakdown of \gmark{} by task.}
    \label{fig:latency_breakdown}
\end{figure}

\begin{figure}[t]
    \centering
    \includegraphics[width=.9\columnwidth]{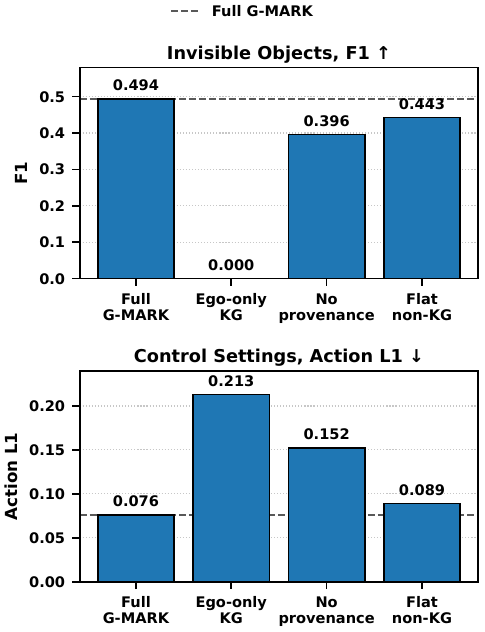}
    \caption{Diagnostic ablations on evidence-sensitive tasks.}
    \label{fig:ablation_diagnostic}
\end{figure}

\textbf{Structured KG reasoning adds minimal CPU-only overhead.}
Fig.~\ref{fig:latency_breakdown} shows representative per-sample CPU runtimes after upstream perception has produced processed scene artifacts. 
We group runtime into scene loading, KG/task solving, and other overhead for graph construction and feature extraction.

The reasoning head is lightweight: task-solver time remains below $1.4$ ms per sample and below $1$ ms for six of the eight task types. 
Current-frame tasks require $10.3$--$14.2$ ms total CPU latency, while temporal motion tasks require $32.4$--$33.4$ ms due to previous-frame context loading. 
Most runtime comes from data access and scene preparation rather than KG reasoning, indicating that the proposed reasoning layer is not the primary bottleneck. Recent VLM-based cooperative driving systems report end-to-end latencies on the order of hundreds of milliseconds~\cite{v2xvlm,v2xrealm}. 
Although not a direct system-level comparison, \gmark{} requires only about $4$--$6\%$ of these reported latencies for current-frame tasks, while preserving millisecond-scale CPU reasoning.

\subsection{Ablation Study}
We conduct construction-level ablations to evaluate which KG evidence components contribute to downstream reasoning. The task heads are kept fixed, while the KG evidence is selectively modified before inference. Fig.~\ref{fig:ablation_diagnostic} summarizes the results for hidden-object discovery and control selection.

\textbf{Partner evidence is essential for hidden-object discovery.}
As shown in Fig.~\ref{fig:ablation_diagnostic}, removing partner observations drops Invisible Objects F1 from $0.494$ to $0.000$. 
This confirms that hidden-object discovery depends on cooperative evidence outside the ego vehicle's direct view; an ego-only representation cannot recover objects visible only to connected vehicles.

\textbf{Provenance affects both reasoning and control.}
Removing provenance reduces Invisible Objects F1 from $0.494$ to $0.396$ and increases Control Settings Action L1 from $0.076$ to $0.152$, showing that source-agent support is useful for downstream decisions.

\textbf{KG-derived structure helps beyond object geometry.}
Replacing the KG with an unstructured object-level representation reduces Invisible Objects F1 from $0.494$ to $0.443$ and increases Control Settings Action L1 from $0.076$ to $0.089$. 
These results suggest that visibility, uncertainty, disagreement, and path-relevance features provide useful structure for evidence-sensitive cooperative reasoning.

\section{Conclusion}
\label{sec:conclusion}

We presented \gmark{}, a grounded multi-agent reasoning framework that maps processed cooperative perception evidence into a provenance-aware KG for object reasoning, prediction, control selection, and trajectory forecasting. On V2V-GoT-QA, \gmark{} improves evidence-sensitive tasks such as occlusion reasoning and hidden-object discovery, while achieving near-parity in future trajectory forecasting with $25.6\times$ lower structured communication and millisecond-scale CPU overhead. These results demonstrate the potential of provenance-aware cooperative KGs as compact, traceable, and communication-efficient reasoning layers for cooperative driving.

\section*{Acknowledgements}
This work has been funded in part by NSF, with award numbers \#2112665, \#2112167, \#2003279, \#2120019, \#2211386, \#2052809, \#1911095 and in part by PRISM and CoCoSys, centers in JUMP 2.0, an SRC program sponsored by DARPA.

\bibliographystyle{IEEEtran}
\bibliography{refs}

@misc{chiu2026v2vgot,
title = {{V2V-GoT}: Vehicle-to-Vehicle Cooperative Autonomous Driving with Multimodal Large Language Models and Graph-of-Thoughts},
author = {Chiu, Hsu-kuang and others},
year = {2025},
eprint = {2509.18053},
archivePrefix= {arXiv},
primaryClass = {cs.RO},
note = {Accepted to ICRA 2026}
}

@article{tian2026kldrive,
  title={KLDrive: Fine-Grained 3D Scene Reasoning for Autonomous Driving based on Knowledge Graph},
  author={Tian, Ye and Zhang, Jingyi and Wang, Zihao and Ren, Xiaoyuan and Yu, Xiaofan and Gungor, Onat and Rosing, Tajana},
  journal={arXiv preprint arXiv:2603.21029},
  year={2026}
}

@misc{chiu2026v2vllm,
title = {{V2V-LLM}: Vehicle-to-Vehicle Cooperative Autonomous Driving with Multi-Modal Large Language Models},
author = {Chiu, Hsu-kuang and others},
year = {2025},
eprint = {2502.09980},
archivePrefix= {arXiv},
primaryClass = {cs.RO},
note = {Accepted to ICRA 2026}
}

@article{kuznietsov2024explainable,
  title={Explainable AI for safe and trustworthy autonomous driving: A systematic review},
  author={Kuznietsov, Anton and others},
  journal={IEEE Transactions on Intelligent Transportation Systems},
  volume={25},
  number={12},
  pages={19342--19364},
  year={2024},
  publisher={IEEE}
}

@inproceedings{xu2022opv2v,
  title={Opv2v: An open benchmark dataset and fusion pipeline for perception with vehicle-to-vehicle communication},
  author={Xu, Runsheng and Xiang, Hao and Xia, Xin and Han, Xu and Li, Jinlong and Ma, Jiaqi},
  booktitle={2022 International Conference on Robotics and Automation (ICRA)},
  pages={2583--2589},
  year={2022},
  organization={IEEE}
}

@article{zhang2023occlusion,
  title={Occlusion-aware planning for autonomous driving with vehicle-to-everything communication},
  author={Zhang, Chi and others},
  journal={IEEE Transactions on Intelligent Vehicles},
  volume={9},
  number={1},
  pages={1229--1242},
  year={2023},
  publisher={IEEE}
}

@inproceedings{v2v4real,
  title={V2V4Real: A Real-World Large-Scale Dataset for Vehicle-to-Vehicle Cooperative Perception},
  author={Xu, Runsheng and others},
  booktitle={Proceedings of the IEEE/CVF Conference on Computer Vision and Pattern Recognition},
  pages={13712--13722},
  year={2023}
}

@inproceedings{v2xvit,
  title={V2X-ViT: Vehicle-to-Everything Cooperative Perception with Vision Transformer},
  author={Xu, Runsheng and Xiang, Hao and Tu, Zhengzhong and Xia, Xin and Yang, Ming-Hsuan and Ma, Jiaqi},
  booktitle={European Conference on Computer Vision},
  pages={107--124},
  year={2022},
  organization={Springer}
}

@inproceedings{cobevt,
  title={CoBEVT: Cooperative Bird's Eye View Semantic Segmentation with Sparse Transformers},
  author={Xu, Runsheng and Tu, Zhengzhong and Xiang, Hao and Shao, Wei and Zhou, Bolei and Ma, Jiaqi},
  booktitle={Conference on Robot Learning},
  pages={989--1000},
  year={2023},
  organization={PMLR}
}

@inproceedings{drivelm,
title = {{DriveLM}: Driving with Graph Visual Question Answering},
author = {Sima, Chonghao and Renz, Katrin and Chitta, Kashyap and Chen, Li and Zhang, Hanxue and Xie, Chengen and Beisswenger, Jens and Luo, Ping and Geiger, Andreas and Li, Hongyang},
booktitle = {Proceedings of the European Conference on Computer Vision (ECCV)},
pages = {256--274},
year = {2024},
publisher = {Springer}
}

@inproceedings{lmdrive,
title = {{LMDrive}: Closed-Loop End-to-End Driving with Large Language Models},
author = {Shao, Hao and others},
booktitle = {Proceedings of the IEEE/CVF Conference on Computer Vision and Pattern Recognition (CVPR)},
year = {2024}
}

@inproceedings{drivevlm,
title = {{DriveVLM}: The Convergence of Autonomous Driving and Large Vision-Language Models},
author = {Tian, Xiaoyu and Gu, Junru and Li, Bailin and Liu, Yicheng and Wang, Yang and Zhao, Zhiyong and Zhan, Kun and Jia, Peng and Lang, Xianpeng and Zhao, Hang},
booktitle = {Proceedings of the Conference on Robot Learning (CoRL)},
year = {2024}
}

@inproceedings{where2comm,
  title={Where2comm: Communication-Efficient Collaborative Perception via Spatial Confidence Maps},
  author={Hu, Yue and others},
  booktitle={Advances in Neural Information Processing Systems},
  volume={35},
  pages={4874--4886},
  year={2022}
}

@inproceedings{langcoop,
  title={LangCoop: Collaborative Driving with Language},
  author={Gao, Xiangbo and others},
  booktitle={Proceedings of the IEEE/CVF Conference on Computer Vision and Pattern Recognition Workshops},
  year={2025}
}

@inproceedings{colmdriver,
  title={CoLMDriver: LLM-based Negotiation Benefits Cooperative Autonomous Driving},
  author={Liu, Chenxi and others},
  booktitle={Proceedings of the IEEE/CVF International Conference on Computer Vision},
  year={2025}
}

@article{drivegpt4,
title = {{DriveGPT4}: Interpretable End-to-End Autonomous Driving via Large Language Model},
author = {Xu, Zhenhua and Zhang, Yujia and Xie, Enze and Zhao, Zhen and Guo, Yong and Wong, Kwan-Yee K. and Li, Zhenguo and Zhao, Hengshuang},
journal = {IEEE Robotics and Automation Letters},
year = {2024},
note = {Also available as arXiv:2310.01412}
}

@book{barshalom1988tracking,
  title={Tracking and Data Association},
  author={Bar-Shalom, Yaakov and Fortmann, Thomas E.},
  year={1988},
  publisher={Academic Press}
}

@article{yurtsever2020survey,
  title={A Survey of Autonomous Driving: Common Practices and Emerging Technologies},
  author={Yurtsever, Ekim and Lambert, Jacob and Carballo, Alexander and Takeda, Kazuya},
  journal={IEEE Access},
  volume={8},
  pages={58443--58469},
  year={2020},
  publisher={IEEE}
}

@article{zhang2021safe,
  title={Safe occlusion-aware autonomous driving via game-theoretic active perception},
  author={Zhang, Zixu and Fisac, Jaime F},
  journal={arXiv preprint arXiv:2105.08169},
  year={2021}
}

@article{v2xvlm,
  title={V2X-VLM: End-to-End V2X Cooperative Autonomous Driving Through Large Vision-Language Models},
  author={You, Junwei and others},
  journal={arXiv preprint arXiv:2408.09251},
  year={2024}
}

@article{v2xrealm,
  title={V2X-REALM: Vision-Language Model-Based Robust End-to-End Cooperative Autonomous Driving with Adaptive Long-Tail Modeling},
  author={You, Junwei and Li, Pei and Jiang, Zhuoyu and Huang, Zilin and Gan, Rui and Shi, Haotian and Ran, Bin},
  journal={arXiv preprint arXiv:2506.21041},
  year={2025}
}

@article{hogan2021knowledge,
  title={Knowledge Graphs},
  author={Hogan, Aidan and others},
  journal={ACM Computing Surveys},
  volume={54},
  number={4},
  pages={1--37},
  year={2021},
  publisher={ACM},
  doi={10.1145/3447772}
}

@article{v2xsim,
  title        = {{V2X-Sim}: Multi-Agent Collaborative Perception Dataset and Benchmark for Autonomous Driving},
  author       = {Li, Yiming and Ma, Dekun and An, Ziyan and Wang, Zixun and Zhong, Yiqi and Chen, Siheng and Feng, Chen},
  journal      = {IEEE Robotics and Automation Letters},
  volume       = {7},
  number       = {4},
  pages        = {10914--10921},
  year         = {2022},
  doi          = {10.1109/LRA.2022.3192802}
}

@misc{w3cprov,
  title        = {{PROV-Overview: An Overview of the PROV Family of Documents}},
  author       = {Groth, Paul and Moreau, Luc},
  howpublished = {W3C Working Group Note},
  year         = {2013},
  month        = apr,
  url          = {https://www.w3.org/TR/prov-overview/},
  note         = {World Wide Web Consortium (W3C)}
}

@book{koller2009probabilistic,
  title={Probabilistic Graphical Models: Principles and Techniques},
  author={Koller, Daphne and Friedman, Nir},
  year={2009},
  publisher={MIT Press}
}

\end{document}